\documentclass[sigconf]{aamas}  

\usepackage{booktabs}

\setcopyright{ifaamas} 
\acmDOI{} 
\acmISBN{} 

\acmConference[AAMAS'18]{\textbf{to appear} as an extended abstract
  paper in the Proc.\@ of the 17th International Conference on
  Autonomous Agents and Multiagent Systems (AAMAS 2018)}{July 10--15,
  2018}{Stockholm, Sweden}{M.~Dastani, G.~Sukthankar, E.~Andr\'{e},
  S.~Koenig (eds.)}

\acmYear{2018} 
\copyrightyear{2018} 
\acmPrice{} 

\usepackage{amsmath}
\usepackage{amssymb}
\usepackage{epsfig}
\usepackage{subfigure}
\usepackage{balance}
\usepackage[ruled,vlined]{algorithm2e}
\usepackage{hyperref}

\newcommand\dop{\textsc{DOP}}
\newcommand\set[1]{\textsc{#1}}

\DeclareMathOperator*{\argmax}{\arg\max}

\begin{document}
\title{\dop: Deep Optimistic Planning with Approximate Value Function Evaluation}



\author{Francesco Riccio}
\affiliation{%
  \institution{Sapienza University of Rome, Department of Computer,
  Control and Management Engineering}
  \city{Via Ariosto 25, Rome} 
  \state{Italy} 
  \postcode{00185}
}
\email{riccio@diag.uniroma1.it}
\author{Roberto Capobianco}
\affiliation{%
  \institution{Sapienza University of Rome, Department of Computer,
  Control and Management Engineering}
  \city{Via Ariosto 25, Rome} 
  \state{Italy} 
  \postcode{00185}
}
\email{capobianco@diag.uniroma1.it}
\author{Daniele Nardi}
\affiliation{%
  \institution{Sapienza University of Rome, Department of Computer,
  Control and Management Engineering}
  \city{Via Ariosto 25, Rome} 
  \state{Italy} 
  \postcode{00185}
}
\email{nardi@diag.uniroma1.it}

\begin{abstract}
  Research on reinforcement learning has demonstrated promising
  results in manifold applications and domains. Still, efficiently
  learning effective robot behaviors is very difficult, due to
  unstructured scenarios, high uncertainties, and large state
  dimensionality (e.g. multi-agent systems or hyper-redundant
  robots). To alleviate this problem, we present \dop{}, a deep
  model-based reinforcement learning algorithm, which exploits action
  values to both (1) guide the exploration of the state space and (2)
  plan effective policies. Specifically, we exploit deep neural
  networks to learn $Q$-functions that are used to attack the
  curse of dimensionality during a Monte-Carlo tree search. Our
  algorithm, in fact, constructs upper confidence bounds on the
  learned value function to select actions optimistically. We
  implement and evaluate \dop{} on different scenarios: (1) a
  cooperative navigation problem, (2) a fetching task for a 7-DOF KUKA
  robot, and (3) a human-robot handover with a humanoid robot (both in
  simulation and real). The obtained results show the effectiveness of
  \dop{} in the chosen applications, where action values drive the
  exploration and reduce the computational demand of the planning
  process while achieving good performance.
\end{abstract}


\maketitle

\section{Introduction}
\label{sec:intro}

Action planning in robotics is a complex task due to
unpredictabilities of the physical world, uncertainties in the
observations, and rapid explosions of the state dimensionality. For
example, hyper-redundant manipulators are typically affected by the
curse of dimensionality problem when planning in large state
spaces. Similarly, in multi-robot collaborative tasks, each robot has
to account for both the state of the environment and other robots'
states. Due to the curse of dimensionality, generalization and policy
generation are time consuming and resource intensive. While deep
learning approaches led to improved generalization capabilities and
major successes in reinforcement learning~\cite{Mnih2015,
  Lillicrap2015, Mnih2016} and robot planning~\cite{Levine2016,
  Levine2016a}, most techniques require huge amounts of data collected
through agent's experience.
\begin{figure}[t!]
  \centering
  \includegraphics[width=0.9\columnwidth]{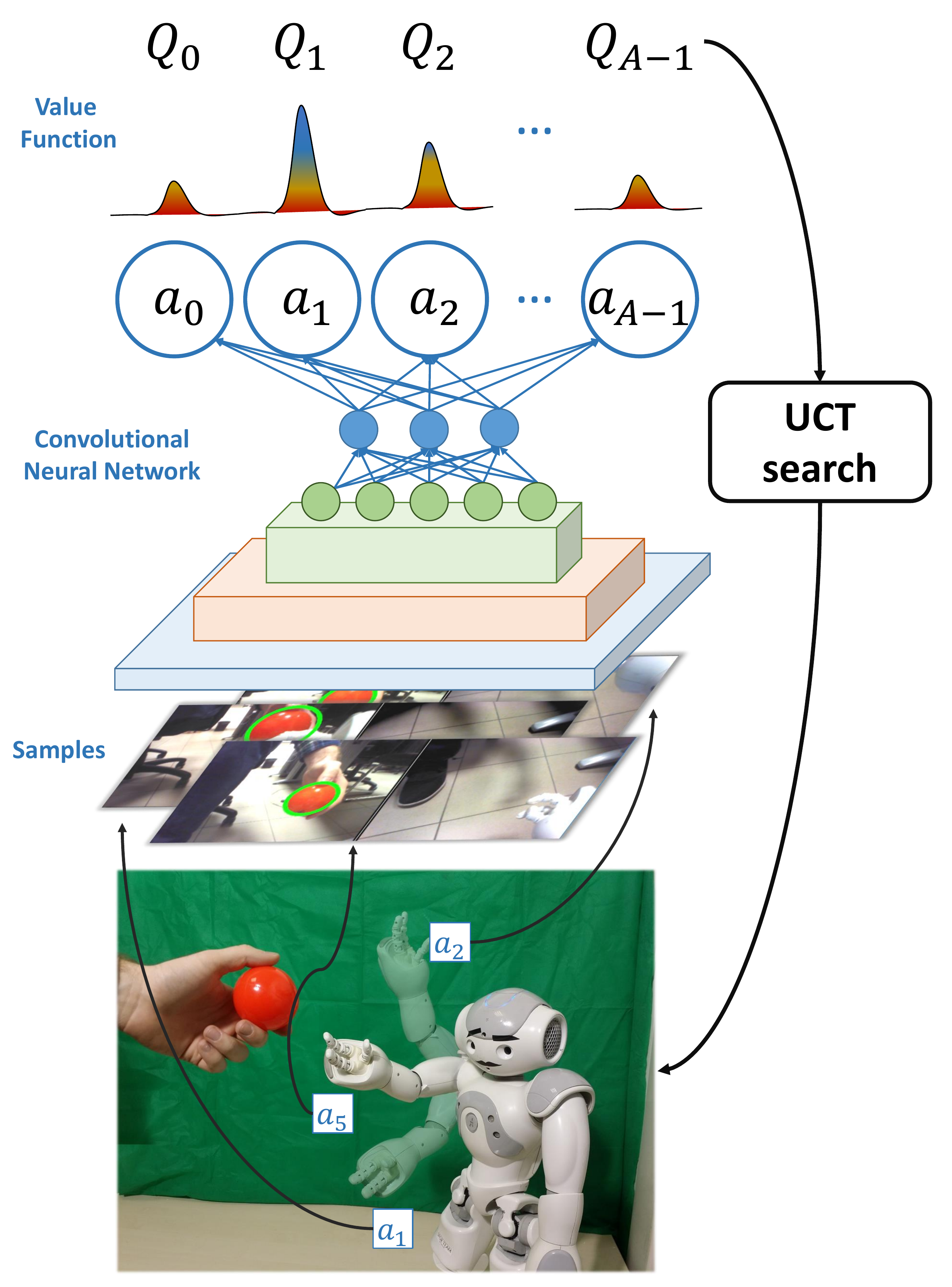}
  \caption{Overview of \dop{}. Action values are learned with a deep
    convolutional neural network over observations collected from
    robot experience. These $Q$-values are then explicitly used at
    planning time to prune actions evaluated during a Monte-Carlo tree
    search.}
  \label{fig:aamas18_intro}
\end{figure}
In robotics, this is often achieved by spawning multiple
simulations~\cite{Rusu2016} in parallel to feed a single neural
network. During simulation, however, the robot explores its huge
search space, with little or no prior knowledge. While encoding prior
information in robot behaviors is often desirable, this is difficult
when using neural networks and it is mostly achieved through imitation
learning~\cite{Calinon2010, Vecerik2017} with little performance
guarantees. To overcome this issue, decision theoretical planning
techniques, such as Monte-Carlo tree search~\cite{Browne2012}, have
been applied in literature. Unfortunately, these methods fail in
generalizing and show limitations in relating similar
states~\cite{Silver2012} (i.e. nodes of the search tree).

As in prior work~\cite{Riccio2018}, we attack the generalization
problem in policy generation by enhancing the Upper Confidence Tree
(UCT) algorithm~\cite{Kocsis2006} -- a variant of Monte-Carlo tree
search -- with an external action-value function approximator, that
selects admissible actions and consequently drives the node-expansion
phase during episode simulation. In particular, in this paper, we
extend the algorithm in~\cite{Riccio2018} to use a more powerful
representation, based on deep learning, that enables better
generalization and supports higher dimensional problems. In fact,
\dop{} (Deep Optimistic Planning), is based on Q-learning and allows
agents to plan complex behaviors in scenarios characterized by
discrete action spaces and large state spaces. As shown in
\figurename~\ref{fig:aamas18_intro}, to model action values, we use a
convolutional neural network (CNN) that is iteratively refined by
aggregating~\cite{Ross2011} samples collected at every
timestep. Specifically, \dop{} generates action policies by running a
Monte Carlo tree search~\cite{Winands2008} and incrementally
collecting new samples that are used to improve a deep $Q$-network
approximating action values.

We aim at demonstrating that \dop{} can efficiently be used to
generalize policies and restrict the search space to support learning
in high-dimensional state spaces. To this end, we address applications
that involve multi and hyper-redundant robots by evaluating \dop{}
over three complex scenarios: a cooperative navigation task of 3
Pioneer robots in a simple environment, a fetching task with a 7-DOF
KUKA arm, and an handover task with a humanoid NAO robot. The
experimental evaluation shows the effectiveness of using deep learning
to represent action values that are exploited in the search process,
resulting in improved generalization capabilities of the planning
algorithm -- that make it suitable also for multi-agent domains. Our
main contributions consist in (1) an extension of prior
work~\cite{Riccio2018} to use deep learning and improve both the
focused exploration and generalization capabilities, and (2) an
extended experimental evaluation of the representation and the
algorithm, that shows the usability of \dop{} in complex
high-dimensional and multi-robot domains.

The reminder of the paper is organized as
follows. Section~\ref{sec:related} reports recent research on (deep)
reinforcement learning and planning, and Section~\ref{sec:q-cp}
describes the \dop{} representation and algorithm. Finally,
Section~\ref{sec:evaluation} describes our experimental setup, as well
as the obtained results, and Section~\ref{sec:conclusion} discusses
final remarks, limitations and future directions.


\section{Related Work}
\label{sec:related}

Recent trends in reinforcement learning have shown improved
generalization capabilities, by exploiting deep learning
techniques. For example, Mnih et al.~\cite{Mnih2015} use a deep
$Q$-network to learn directly from high-dimensional visual sensory
inputs on Atari 2600 games. Silver et al.~\cite{Silver2016a,
  Silver2017} use deep value networks and policy networks to
respectively evaluate board positions and select moves to achieve
superhuman performance in Go. In~\cite{Mnih2016}, instead, the authors
execute multiple agents in parallel, on several instances of the same
environment, to learn a variety of tasks using actor-critic with
asynchronous gradient descent. Similar advancements have been shown in
the robotics domain. For instance, Levine et al.~\cite{Levine2016}
represent policies through deep convolutional neural networks, and
train them using a partially observed guided policy search method on
real-world manipulation tasks. Moreover, Rusu et al.\cite{Rusu2016}
use deep learning in simulation, and propose progressive networks to
bridge the reality gap and transfer learned policies from simulation
to the real world. Unfortunately, planning and learning with deep
networks is computationally demanding (i.e., requires a huge number of
heavy simulations). For this reason, Weber et al.~\cite{Weber2017}
introduce I2As to exploit outcome of virtual policy executions, in the
Sokoban domain, learned with a deep network.

During simulation, the robot uninformedly explores its search space,
and attempts to find an optimal policy. To facilitate this task,
several approaches initialize robot policies with reasonable behaviors
collected from human experts (e.g., through the aid of expert
demonstrations~\cite{Mnih2015, Sun2017}). For example,
in~\cite{Vecerik2017} the authors rely on a Deep Deterministic Policy
Gradient approach to exploit expert demonstration in object insertion
tasks. These methods, however, generally require a considerable number
of training examples, and are not easily applicable to complex
domains, such as multi-agent scenarios and highly redundant
robots. While traditional decision theoretic planning methods, such as
Monte-Carlo tree search~\cite{Kocsis2006, Browne2012}, enable easier
injection of prior knowledge in generated behaviors, they do not
provide sufficient generalization capabilities, that are required in
common robotics problems, where large portions of the state space are
rarely or never explored.

In this paper, we address generalization at learning time by
introducing \dop, an iterative algorithm for policy generation that
makes use of deep $Q$-networks to drive a focused exploration. \dop{}
relies on previous work~\cite{Riccio2018} and extends it with a
different representation to obtain improved generalization
capabilities over higher dimensional problems. In particular, we
approximate the $Q$ function using $Q$-learning with a deep
convolutional neural network. Similar to $Q$-CP~\cite{Riccio2018} and
TD-search~\cite{Silver2012}, we aim at reducing the variance of value
estimates during the search procedure by means of temporal difference
learning.  However, as in~\cite{Riccio2018}, \dop{} extends TD-search
by constructing upper confidence bounds on the value function, and by
selecting optimistically with respect to those, instead of performing
$\epsilon$-greedy exploration. Thanks to the generalization
capabilities of deep networks, \dop{} improves over $Q$-CP both in
terms of policy and exploration. Our representation, in fact, not only
enables the algorithm to explore more informative portions of the
search space~\cite{Gelly2011, James2017}, but also is able to
generalize better among them with positive effects on the overall
policy and search space expansion.


\section{\dop}
\label{sec:q-cp}

We describe \dop{} by adopting the standard Markov Decision Process (MDP)
notation, in which the decision-making problem is represented as a
tuple
\begin{align}
  MDP = (\set{S}, \set{A}, \mathcal{T}, R, \gamma), \nonumber
\end{align}
where $\set{S}$ is the set of states of the environment, $\set{A}$
represents the set of discrete actions available to the agent,
$\mathcal{T}: \set{S} \times \set{A} \times \set{S} \rightarrow [0,
1]$
is the stochastic transition function that models the probabilities of
transitioning from state $s \in \set{S}$ to $s' \in \set{S}$ when an
action is executed, $R: \set{S} \times \set{A} \rightarrow \mathbb{R}$
is the reward function. In this setting, actions are chosen according
to a policy $\pi(a|s)$ (or more simply $\pi(s)$), that maps states to
actions. Given an MDP, the goal of the agent consists in finding a
policy $\pi(s)$ that maximizes its future reward with a discount
factor $\gamma \in [0, 1)$.

Under this framework, \dop{} builds on~\cite{Riccio2018} and
iteratively evolves by (1) running a Monte-Carlo tree search, where
admissible actions are selected through $Q$-value estimates learned
through a deep neural network and (2) incrementally collecting new
samples that are used to improve $Q$-value estimates. In this section,
we first describe our representation for the $Q$ function; then, we
present an explanation of the algorithm.


\subsection{Representation}
\label{sec:representation}

\begin{figure}[t!]
  \centering
  \includegraphics[width=\columnwidth]{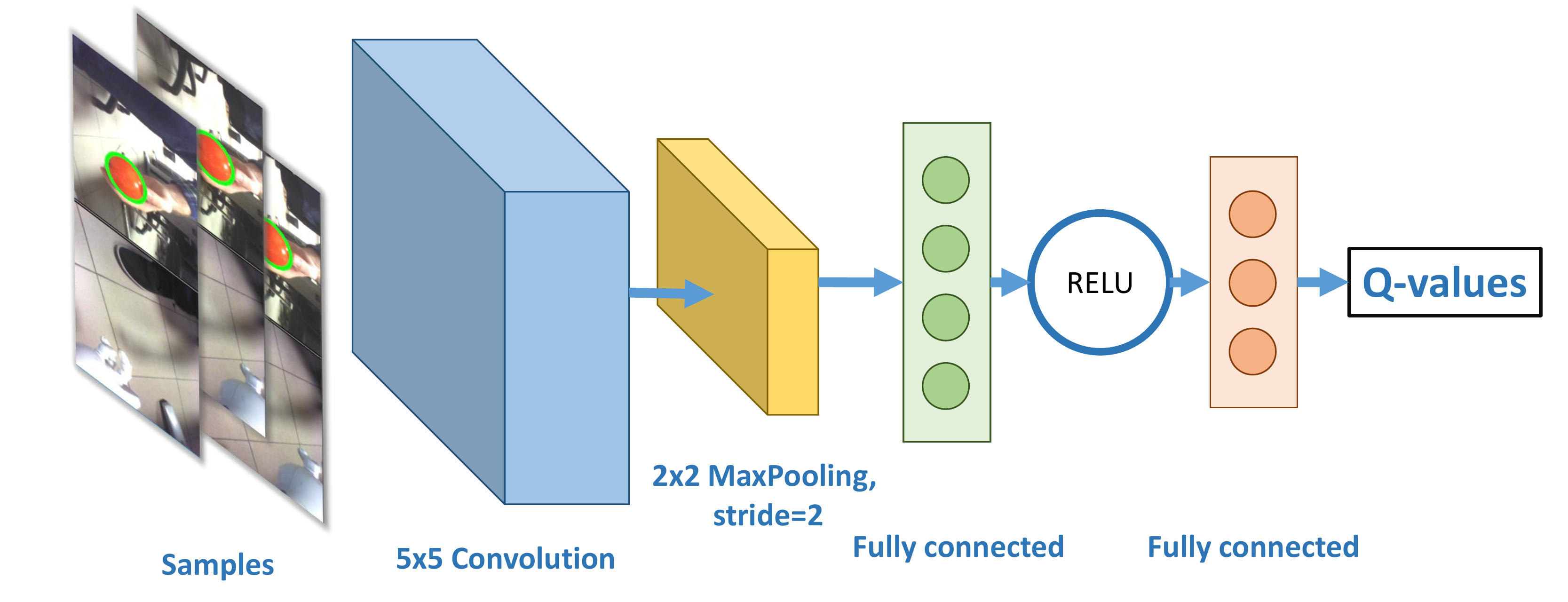}
  \label{fig:dop_net}
  \caption{Deep convolutional neural network adopted in \dop. The
    network is implemented within the Caffe2 environment.}
\end{figure}
As in previous literature~\cite{Mnih2015, Mnih2016}, we choose to
change the representation adopted in~\cite{Riccio2018} and approximate
the action values using a deep neural network. In particular, we adopt
a deep neural network (implemented in
Caffe2\footnote{\url{https://caffe2.ai}}), composed by a convolutional
layer, followed by a max pooling operator and two fully connected
layers, one with ReLU activations, the other linear. The input of the
network is an image capturing the state of the environment, and its
output is the $Q$ value for each action. In our work, we always assume
one single frame to be sufficient to fully represent the state $s$ at
each timestep, and we store in our dataset transitions $x =
(s_{t},a_{t}, s_{t+1}, r_{t+1})$.

Differently from DQN, we do not explicitly make use of an experience
replay buffer, and we replace it with a data
aggregation~\cite{Ross2011} procedure, where all the transitions are
iteratively collected, aggregated and used at training
time. Specifically, at each iteration $i$ of \dop{} we collect a
dataset $\set{D}^{i} = \{x\}$ of transitions experienced by the agent,
and we aggregate it into
$\set{D}^{0:i} = \{\cup\set{D}^{d}|d=0 \dots i\}$, that is used for
learning. In practice, this results in a very similar mechanism that
enables data decorrelation and facilitates learning. In fact, the
network is always re-trained using mini-batches of samples randomly
selected within $\set{D}^{0:i}$, until the dataset is exhaustively
used. Hence, each mini-batch is less affected by the non-i.i.d
structure of the dataset, and this procedure closely resembles the
more standard replay mechanism. 

The aggregated dataset is finally used to minimize the $\ell_2$-loss:
\begin{align}
  \label{eq:q_rule_update}
  \ell_2(r_{t+1} + \gamma\max_{a'}Q_{\theta}(s_{t+1},a'),
  Q_{\theta}(s_{t},a_{t})),
\end{align}
where $\theta$ are the current parameters of the network. The
optimization is performed using an Adam~\cite{Kingma2014} optimizer
with a learning rate $\alpha$ selected according to the task.


\subsection{Algorithm}
\label{sec:q-cp-algo}

\dop{} builds on $Q$-CP~\cite{Riccio2018}, which is an iterative
algorithm that, at each iteration $i$, (1) generates a new policy
$\pi_i$ which improves $\pi_{i-1}$ and (2) learns action values that
are used at planning time to reduce the search space. In this section,
we briefly describe the algorithm by adapting its discussion to the
representation adopted in \dop. Specifically, at every iteration the
agent executes an UCT search, where admissible actions are selected
through $Q$ value estimates, and incrementally collects new samples
that are used to improve $Q$ value estimates as described in previous
section.

More in detail (see Algorithm~\ref{alg:q-cp}), \dop{} takes as input
an initial policy $\pi^0$ and, at each iteration $i = 1 \dots I$,
evolves as follows:
\begin{enumerate}
\item the agent follows its policy $\pi^{i-1}$ for $T$ timesteps,
  generating a set of $T$ states $\{s_t\}$;
\item for each generated state $s_t$, the agent runs a modified UCT
  search~\cite{Kocsis2006} with depth $H$. Specifically, at each
  $h = 1 \dots H$, the search algorithm
  \begin{enumerate}
  \item evaluates a subset of ``admissible'' actions
    $\tilde{\set{A}} \subseteq \set{A}$ in $s_{t+(h-1)}$, that are
    determined according to $Q_{\theta}(s_{t+(h-1)},a)$. In
    particular, differently from vanilla UCT, we only allow actions
    $a$ such that
    \begin{align}
      \begin{split}
        Q_{\theta}(s_{t+(h-1)},a) >=~
        \lambda\max_{a}Q_{\theta}(s_{t+(h-1)},a) +
        \epsilon_{\tilde{A}}
      \end{split}
    \end{align}
    \noindent where $\lambda$ is typically initialized to $0.5$ and
    increases with the number of iterations $i$. Through
    $\epsilon_{\tilde{A}}$, a certain amount of exploration is
    guaranteed;
  \item selects and executes the best action
    $a^{*}_{h} \in \tilde{\set{A}}$ according to
    \begin{align}
      \small
      \begin{split}
        \label{eq:uct_max_action}
        e &= C \cdot
        \sqrt{\frac{\log(\sum_{a}{\eta(s_{t+(h-1)},a)})}{\eta(s_{t+(h-1)},a)}} \\
        a^{*}_{h} &= \argmax_{a} Q_{\theta}(s_{t+(h-1)}, a) + e,
      \end{split}
    \end{align}
    where $C$ is a constant that multiplies and controls the
    exploration term $e$, and $\eta(s_{t+(h-1)},a)$ is the number of
    occurrences of $a$ in $s_{t+(h-1)}$. To use \dop{} on robotic
    applications with images as state representations, we define a
    comparison operator that introduces a discretization by computing
    whether the difference of two states is smaller than a given
    threshold $\xi$;
  \item runs $M$ simulations (or roll-outs), by executing an
    $\epsilon$-greedy policy based on $\pi^{i-1}$ until a terminal
    state is reached.
  \end{enumerate}
  \dop{} uses UCT as an expert and collects, a dataset $\set{D}^{i}$
  of $H$ transitions experienced in simulation.
\item $\set{D}^{i}$ is aggregated into
  $\set{D}^{0:i} = \set{D}^{i} \cup
  \set{D}^{0:i-1}$~\cite{Ross2011,Ross2014},
  and the dataset is used to perform updates of the $Q$-network, as
  illustrated in previous section.
\item once $Q_{\theta}$ is updated, the policy is generated as to
  maximize the action values:
  $\pi^{i}(s) = \argmax_{a} Q_{\theta}(s,a)$.
\end{enumerate}

\begin{algorithm}[!t]
  \small
  \DontPrintSemicolon \SetAlgoLined 
  \SetNlSty{}{}{)}
  \SetAlgoNlRelativeSize{0}

  \KwData{$I$ the number of iterations; $\Delta$
    initial state distribution; $H$ UCT horizon; $T$ policy execution
    timesteps, $\lambda_0$ initial max. $Q$ threshold multiplier for
    admissible actions, $\epsilon_{\tilde{A}}$ probability for
    random admissible actions, $\alpha$ learning rate, $\gamma$
    discount factor.}

  \KwIn{$\pi^0$ initial policy of the agent.}

  \KwOut{$\pi^{I}$ policy learned after I iterations.}
  
  \Begin{
    \For{$i = 1$ \KwTo $I$} {
      $s_0 \leftarrow$ random state from $\Delta$.\;
      \For{$t = 1$ \KwTo $T$} {
        \nl Get state $s_t$ by executing $\pi^{i-1}(s_{t-1})$.\;

        \nl $\set{D}^{i} \leftarrow$ UCT$_{\dop}$($s_t,
        \lambda_0, \epsilon_{\tilde{A}}$).\;
        \nl $\set{D}^{0:i} \leftarrow \set{D}^{i} \cup
        \set{D}^{0:i-1}$.\;
        $Q_{\theta}$.UPDATE($\set{D}^{0:i}, \alpha, \gamma$).\;
        \nl $\pi^{i}(s) \leftarrow \argmax_{a} Q_{\theta}(s,a)$.\;
      }
    }

    \BlankLine
    \Return{$\pi^{I}$}\; 
  }
  \caption{\dop}
  \label{alg:q-cp}
\end{algorithm}


\begin{figure*}[t!]
  \subfigure[Cooperative Navigation] {
    \includegraphics[width=0.3\textwidth]{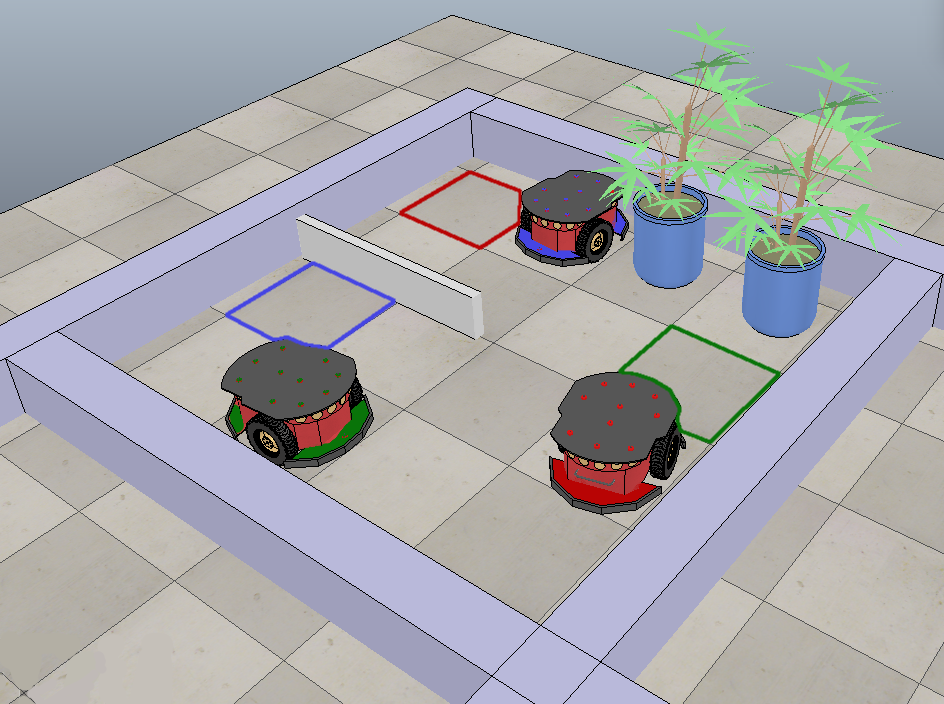}
    \label{fig:simple_intro}
  }
  \subfigure[Handover task] {
    \includegraphics[width=0.3\textwidth]{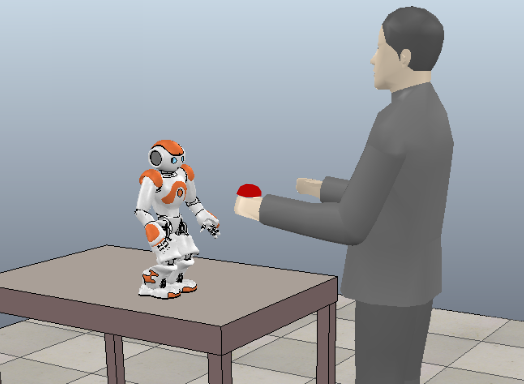}
    \label{fig:nao_intro}
  }
  \subfigure[Fetching task] {
    \includegraphics[width=0.3\textwidth]{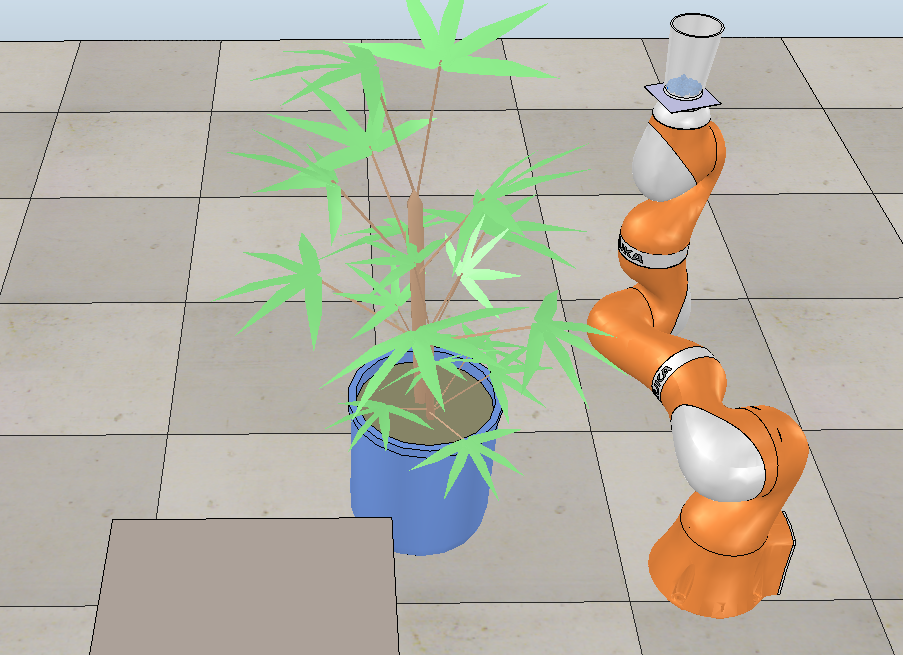}
    \label{fig:lbriiwa_intro}
  }
  \caption{We evaluate \dop{} on three scenarios with different
    robots. Left: cooperative navigation scenario; middle:
    handover with the NAO robot; right: the fetching task with the
    7-DOF KUKA arm.}
  \label{fig:experiment-tasks}
\end{figure*}

\section{Experimental Evaluation}
\label{sec:evaluation}

One of the main contributions of this paper consists in an extended
experimental evaluation of the representation and the algorithm, that
shows the usability of \dop{} in complex high-dimensional and
multi-agent domains. In particular, we evaluate \dop{} against
multiple algorithms over a set of 3 different tasks, as shown in
\figurename~\ref{fig:experiment-tasks}. Specifically, we run our
experiments on three high-dimensional problems, where either a robot
is hyper-redundant, or multiple robots are involved: (1) a
\textit{cooperative navigation} application, where three robots have
to coordinate in order to find the minimum path to their respective
targets in a simple environment; (2) a \textit{fetching task}, where a
7-DOF robotic arm has to fetch an object while avoiding an obstacle in
the environment; and (3) a \textit{handover task} among a NAO
(humanoid) robot and a human.  

\noindent
To report our results, we compare against DQN~\cite{Mnih2015},
TD-search~\cite{Silver2012} and both a \textit{vanilla-UCT} and
\textit{random-UCT} implementations. We refer to \textit{vanilla-UCT}
as the standard UCT algorithm that, at each iteration, expands every
possible action in $A_j$, for every agent $j$. \textit{Random-UCT},
instead, is a naive algorithm where at each step of the Monte-Carlo
search one action is randomly expanded. Moreover, in the
\textit{handover} (\figurename~\ref{fig:nao_intro}) scenario, we
compare the performance of \dop{} against the base $Q$-CP
algorithm~\cite{Riccio2018} and, we present the cumulative reward
obtained by transferring the learned policy on a real NAO.


\subsection{Experimental Setup}
\label{subsec:setup}

Experiments have been conducted using the V-REP simulator running on a
single Intel Core i7-5700HQ core, with CPU@2.70GHz and 16GB of RAM.
For all the scenarios, unless otherwise specified, the algorithm was
configured with the same meta-parameters. The UCT horizon is set to
$H = 4$ to trade-off between usability and performance of the search
algorithm; the number of roll-outs is set to be $M = 3$, while
admissible actions are evaluated with an initial $\lambda = 0.5$, and
$\epsilon_{\tilde{{A}}} = 0.3$, guaranteeing good amounts of
exploration. The $C$ constant in Eq.~\ref{eq:uct_max_action} is set to
$0.7$. The learning rate $\alpha$ is set to $0.15$, while the discount
factor $\gamma$ is equal to $0.8$. The state space, the set of actions
and the reward functions are finally chosen and implemented depending
on the robots and applications. In each of the proposed applications,
stochastic actions are obtained by randomizing their outcomes with a
$5$\% probability. We evaluate the cumulative reward obtained during
different executions of \dop{} against the number of explored states
and iterations of the algorithms. In our environments, we adopt a
\textit{shaped} reward function, thus providing a reward to the agent
at each of the visited states.

\subsection{Cooperative Navigation}
\label{sec:cooperative_navigation}

\begin{figure}[t!]
  \centering
  \subfigure[Rewards] {
    \includegraphics[width=\columnwidth]{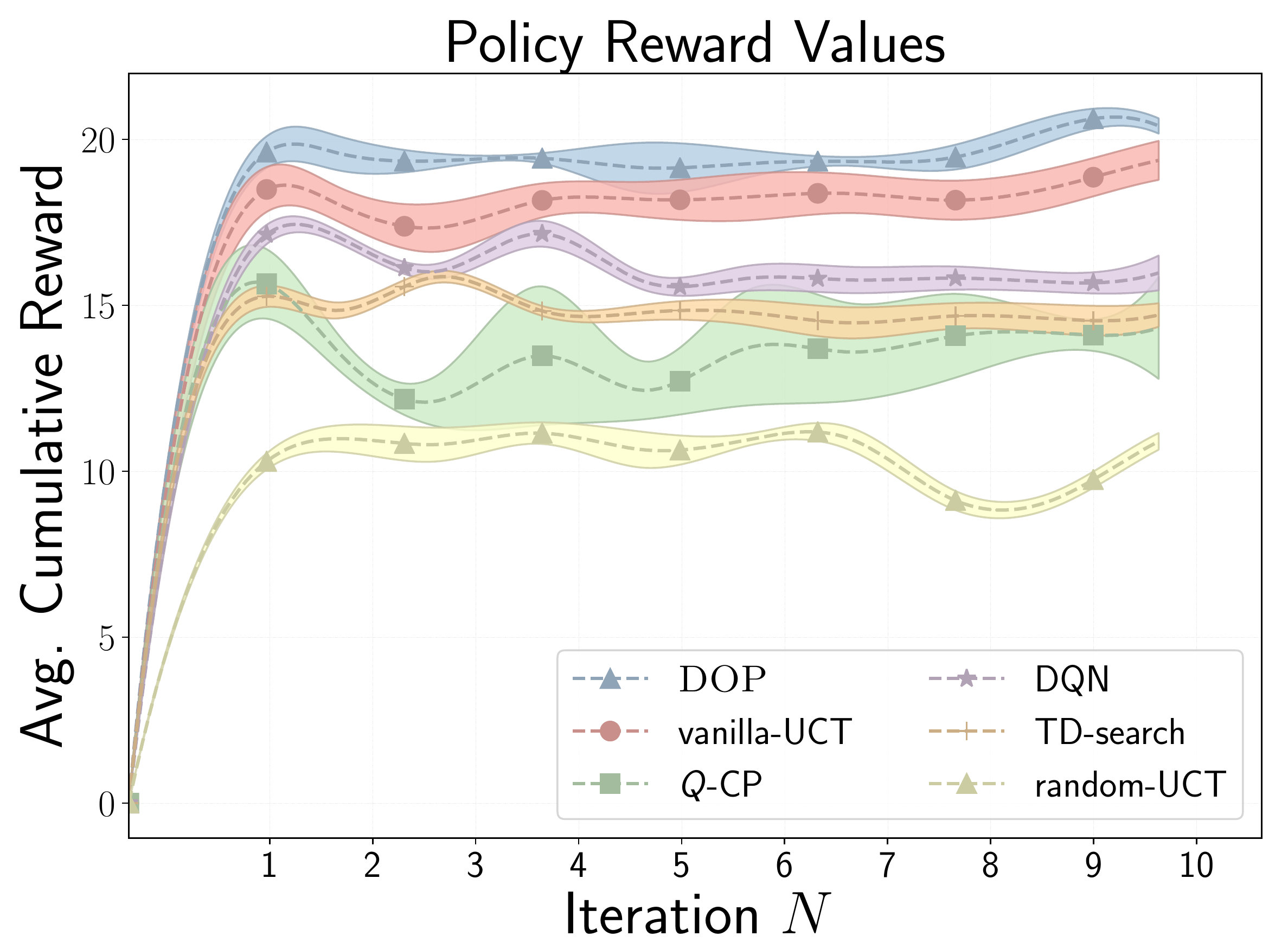}
    \label{fig:simple_rewards}
  }
  \subfigure[States] {
    \includegraphics[width=\columnwidth]{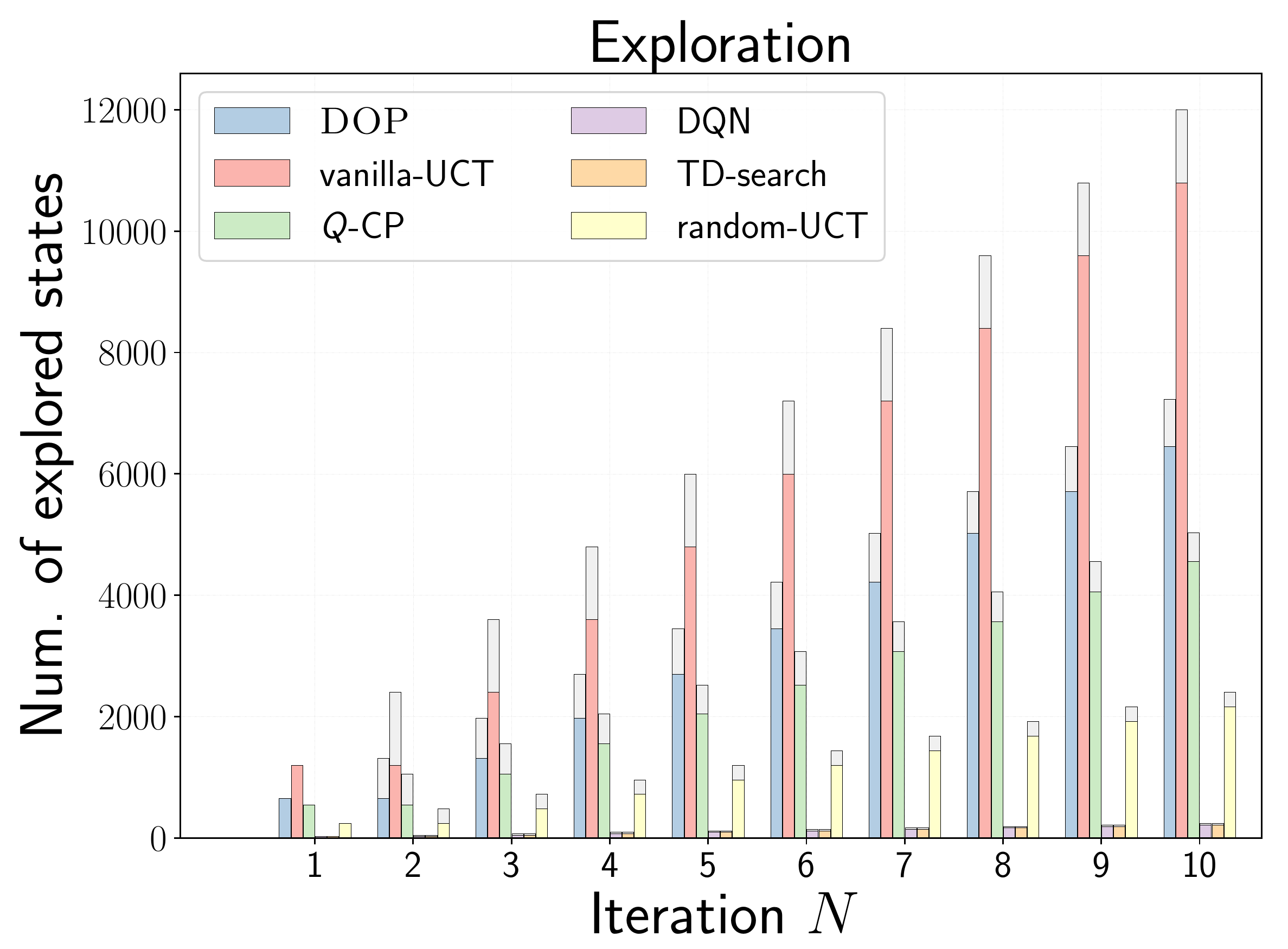}
    \label{fig:simple_states}
  }
  \caption{Average cumulative reward (a) and number of explored states
    (b) obtained by \dop, DQN, TD-search, \textit{random-UCT} and
    \textit{vanilla-UCT} in 10 iterations in the cooperative
    navigation scenario. For each of them, the cumulative reward is
    averaged over 10 runs. The averaged reward is represented as a
    dotted line while, the line width represent its standard
    deviation. At each iteration, the number of explored states is
    represented as a bar, where the height indicates the total number
    of visited states, and the top gray bar highlights the amount of
    states added during the particular iteration $i$.}
  \label{fig:simple}
\end{figure}
In this scenario, three robots have to perform a cooperative
navigation task in a simple world composed by 16 reachable squares
distributed on a 4x4 grid (see
\figurename~\ref{fig:simple_intro}). The goal is assigned individually
to each robot and consists in finding the minimum collision-free path
to target square matching the their color (highlighted in the
figure). The state is represented through an image collected from the
top by an overlooking camera (i.e., all the robots are visible), while
the set of discrete actions is composed by $A = \langle$ \texttt{noop,
  up, down, right, left} $\rangle$. The reward function is normalized
between $[0, 1]$ and is shaped to be inversely proportional to the sum
of the minimum path steps from the robot positions and
targets. \figurename~\ref{fig:simple} shows the results obtained by
\dop{} against $Q$-CP~\cite{Riccio2018}, TD-search, DQN,
\textit{random-UCT} and
\textit{vanilla-UCT}. \figurename~\ref{fig:simple_rewards} reports the
average cumulative reward and standard deviation (line width) over 10
iterations -- averaged over 10 runs -- for each of the implemented
algorithms. \figurename~\ref{fig:simple_states}, reports the number of
states explored per iteration. Bars represent the number of states
explored until the $i$th iteration, and the top gray bar highlights
the amount of states expanded during iteration $i$ against the number
of states explored until $i-1$. The results show different behaviors
for each of the algorithm: \textit{random-UCT} reports a low number of
explored states -- since it only expands one random action at each UCT
iteration -- but, as expected, it performs poorly. Similarly,
TD-search and DQN show a good trade-off between number explored states
and cumulative reward, however, they are not able to generate
competitive policies with few iterations and a reduced number of
training samples. On the other hand, if we consider
\textit{vanilla-UCT} and \dop{} the algorithms report comparable
cumulative rewards. However, our algorithm is able to generate a
competitive policy with a reduced number of explored states --
typically half the number of \textit{vanilla-UCT} ($\sim$45\%
lower). $Q$-CP obtains a lower cumulative reward while visiting a
slightly reduced number of states with respect to \dop. In fact, its
reduced representational power does not enable the algorithm to
immediately generalize informative states that need to be explored to
obtain a better policy. Finally, it is worth highlighting that, even
if both DQN and TD-search present a good trade-off between number of
explored states and performance, they are not able to complete the
task since initial iterations.

\subsection{Fetching Task}

\begin{figure}[t!]
  \centering
  \subfigure[Rewards] {
    \includegraphics[width=\columnwidth]{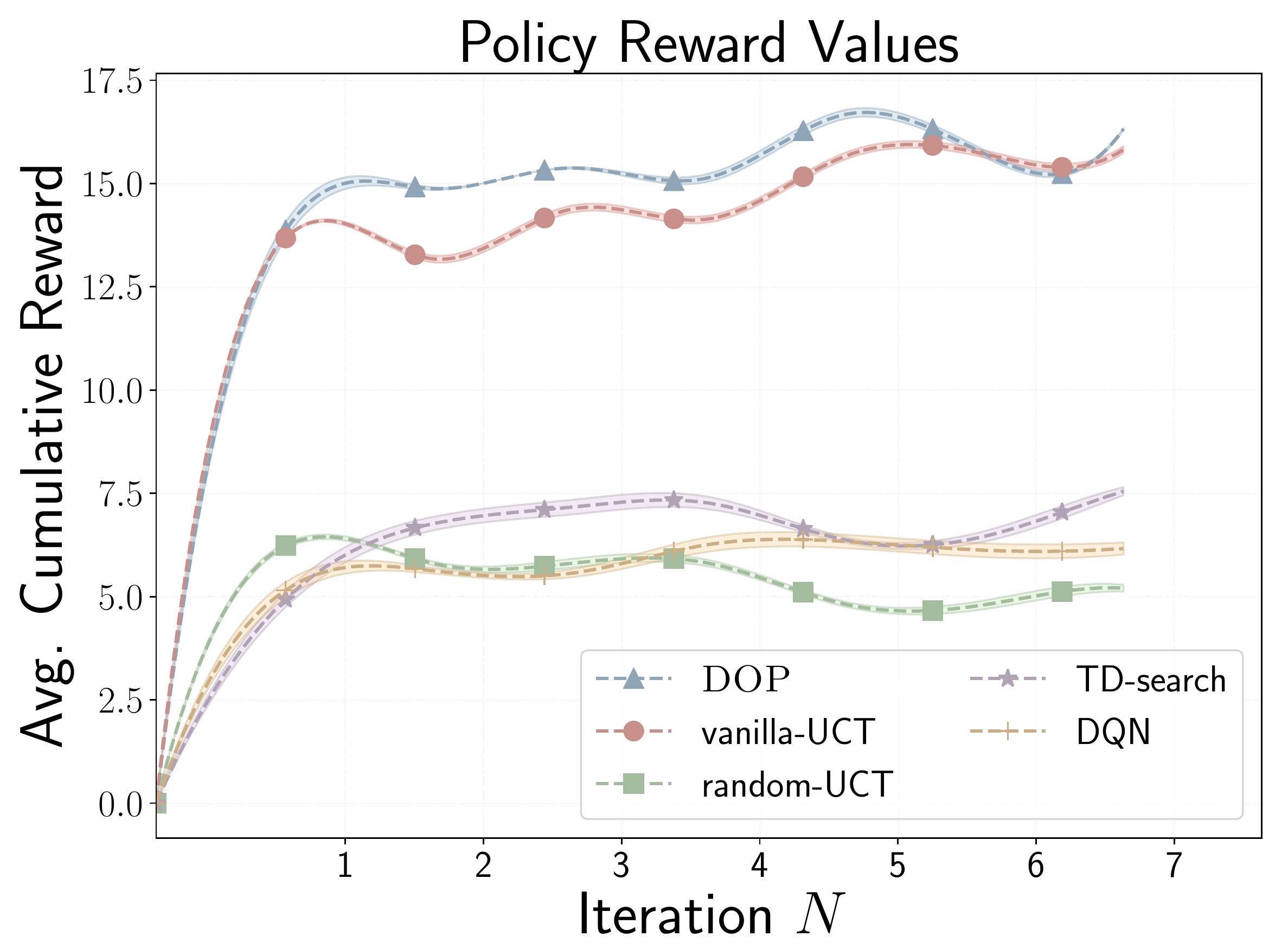}
    \label{fig:lbriiwa_rewards}
  }
  \subfigure[States] {
    \includegraphics[width=\columnwidth]{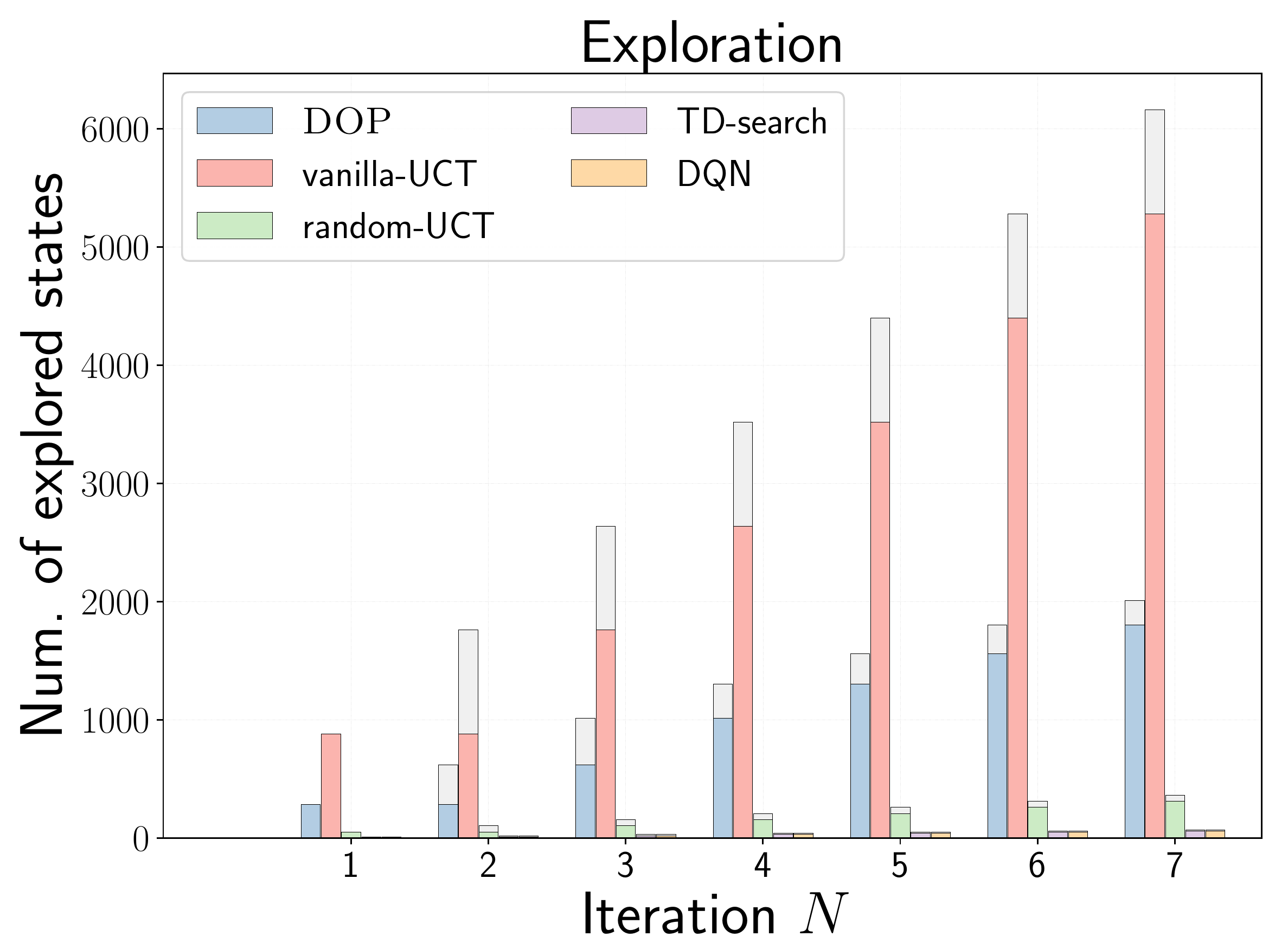}
    \label{fig:lbriiwa_states}
  }
  \caption{Average cumulative reward and number of explored states
    obtained by \dop{}, DQN, TD-search, \textit{random-UCT} and
    \textit{vanilla-UCT} in 7 iterations in the Kuka fetching
    scenario. For each of them, the reward is averaged over 10 runs.
  }
  \label{fig:lbriiwa}
\end{figure}
Here, the 7-DOF KUKA lightweight arm has to learn to fetch an object
(e.g. a glass, \figurename~\ref{fig:lbriiwa_intro}) while avoiding an
obstacle (a plant). In this scenario, the state space is again
represented through an image collected by an overlooking camera, and
the discretization of the state space is realized as described in
Section~\ref{sec:q-cp-algo}. The robot can perform 10 actions:
$A = \langle$\texttt{arm-up}, \texttt{arm-down}, \texttt{arm-forward},
\texttt{arm-backward},
\texttt{arm-right}, \texttt{arm-left}, \texttt{pitch-turn-left},
\texttt{pitch-\\turn-right}, \texttt{yaw-turn-left},
\texttt{yaw-turn-left}$\rangle$.  Rotations on the roll angle have
been removed as they do not influence the desired orientation of the
fetched object. The reward function is in $[0, 1]$ and it is computed
as a weighted sum of four components: the first is inversely
proportional to the Euclidean distance of the end-effector to the
target, the second it proportional to the distance to the virtual
center of the obstacle, the third and the fourth are inversely
proportional to the pitch and yaw angle respectively. In this way the
reward function promotes states that are near the target, far from the
obstacle, and with the end-effector oriented upwards. This is
implemented to succeed in the fetching task of objects that are to be
carried with a preferred orientation (e.g. a glass full of water). As
for the cooperative navigation scenario, \figurename~\ref{fig:lbriiwa}
shows the performance of \dop{}, DQN, TD-search, \textit{random-UCT}
and \textit{vanilla-UCT}. This scenario is key to highlight that our
contribution is more suitable and practical in robotic tasks with
large state space. In fact, since first iterations and with a reduced
set of training samples, \dop{} is able to outperform other algorithms
that need a huge training set to learn competitive policies,
e.g. DQN. Still, \textit{vanilla-UCT} shows comparable rewards, but
the number of explored states for this algorithm is $\sim$65\% larger
than \dop{}.

\subsection{Human-Robot Handover}
\begin{figure}[t!]
  \centering
  \subfigure[Rewards] {
    \includegraphics[width=\columnwidth]{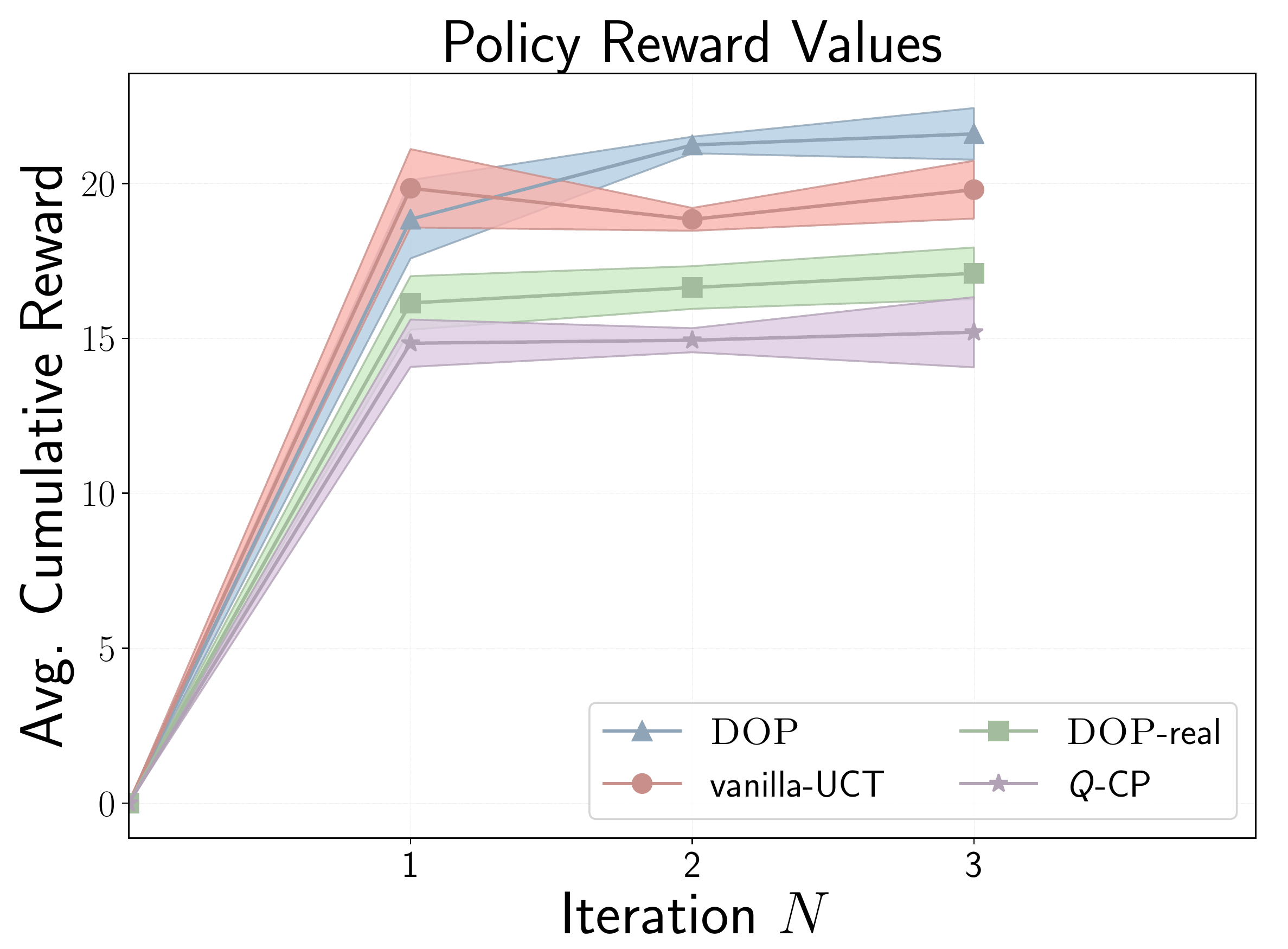}
    \label{fig:nao_rewards}
  }
  \subfigure[States] {
    \includegraphics[width=\columnwidth]{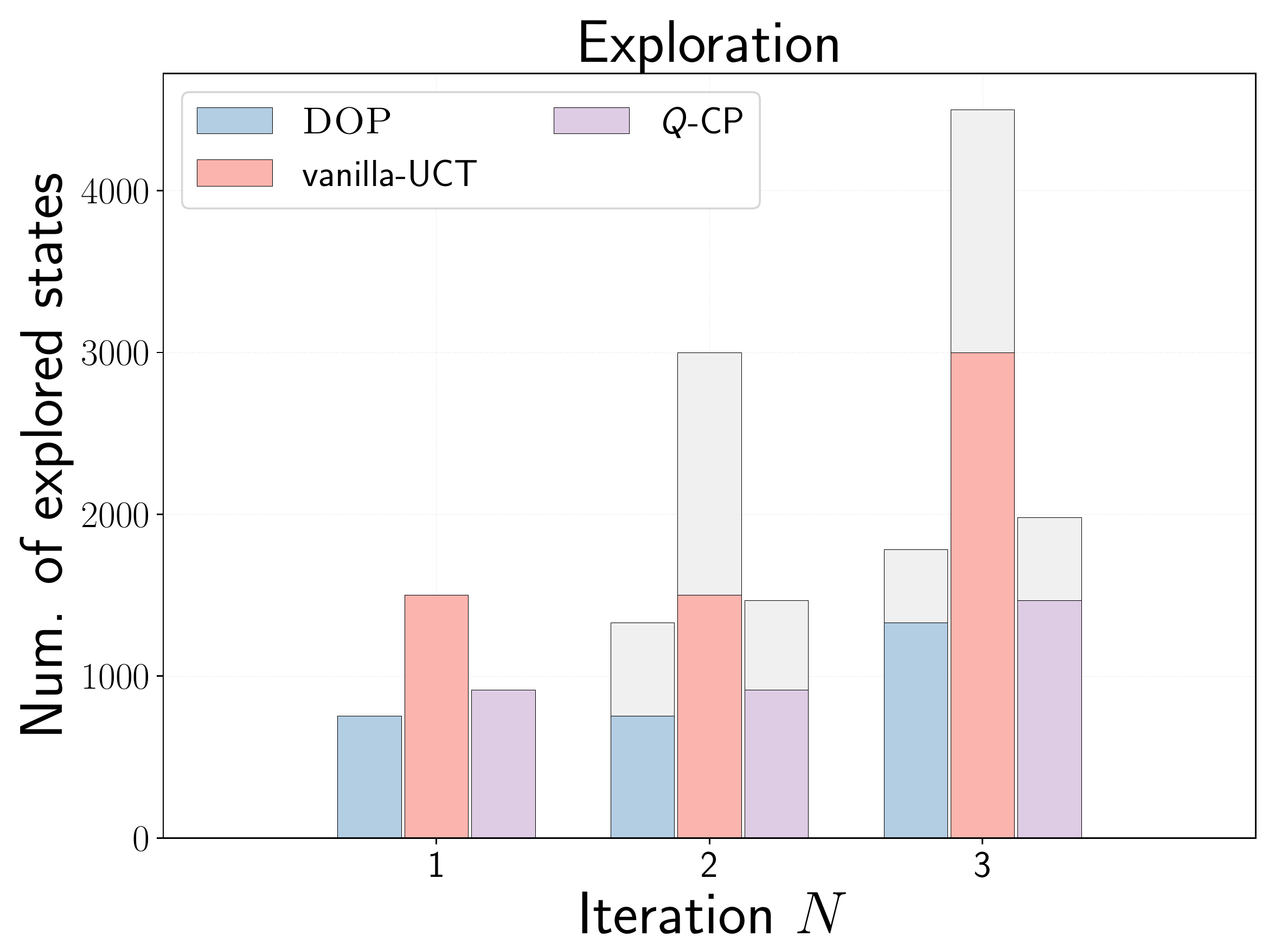}
    \label{fig:nao_states}
  }
  \caption{Average cumulative reward and number of explored states
    obtained by \dop{}, \textit{vanilla-UCT} and $Q$-CP in 3 iterations
    in the handover scenario. \dop{}-real reports reward values by
    transferring the \dop{} learned policy on a real robot. For each
    of them, the reward is averaged over 10 runs.}
  \label{fig:nao}
\end{figure}

The last scenario is characterized by two agents performing a
human-robot interaction. The robot has to complete an handover by
receiving an object which is hold by a human. In this setting, the
state is represented through the images collected by the cameras of
the robot. Moreover, the agent can perform a set of 25 actions:
$A = \langle$\texttt{body-noop}, \texttt{body-forward},
\texttt{body-backward}, \texttt{body-turn-left,\\}
\texttt{body-turn-right},
\texttt{head-up}, \texttt{head-down}, \texttt{head-right},
\texttt{head-left}, \texttt{right-arm-up}, \texttt{right-arm-down},
\texttt{right-arm-left}, \texttt{right-arm-\\right},
\texttt{right-arm-forward}, \texttt{right-arm-backward},
\texttt{left-arm-up}, \texttt{left-arm-down}, \texttt{left-arm-left},
\texttt{left-arm-right}, \texttt{left-arm-fo-\\rward},
\texttt{left-arm-backward}, \texttt{open-right-hand},
\texttt{open-left-hand}, \texttt{close-right-hand},
\texttt{close-left-hand}$\rangle$.

The (shaped) reward function is in $[0, 1]$ and it is implemented as a
weighted sum of 6 components: the first component is inversely
proportional to the distance between the robot and the handed object,
the second and third are inversely proportional to the distance from
the object to the left and right hand respectively, the fourth
component is inversely proportional to the distance between the ball
and the center of the robot camera computed directly in the image
frame, and the fifth and sixth components model the desired status of
the hands, promoting states that with open hands near the handed
object. For this task, we choose a learning rate $\alpha = 0.2$. 

The purpose of this experimental evaluation is to highlight both the
quality of the learned policy when transferred in a real setting and
the improvements of our representation with respect to $Q$-CP. Hence,
we compare the results obtained by \dop{} against $Q$-CP and
\textit{vanilla-UCT}. As in previous scenarios, the two plots
represent the obtained cumulative average reward and the number of
explored states (\figurename~\ref{fig:nao}). As expected, \dop{}
preserves the same improvements over the \textit{vanilla-UCT} and,
also in this case, our algorithm generates an effective policy with a
remarkably reduced number of training examples. Differently, when
comparing \dop{} against $Q$-CP we notice a similar number of explored
states while \dop{} obtains higher cumulative rewards. This shows the
improved generalization capabilities of our representation, that is
able to significantly improve performance while preserving sample
complexity of the original algorithm. Finally, \dop{}-real shows the
rewards obtained by transferring and rolling out the policy in real
settings. In this case, the reward values show that, even though the
robot does not perform as well as in simulation, it is still able to
complete the task being robust to the noise of the real-world
deployment. In fact, we consider the reduced cumulative reward values
to be mostly due to noise in both the perception pipeline and motor
encoders of the real NAO robot.

\section{Conclusion}
\label{sec:conclusion}

In this paper we introduced \dop, an iterative algorithm that uses
action values learned through a deep $Q$-network to guide and reduce
the exploration of the state space in different high-dimensional
scenarios. Our key contribution consists in an extension of
$Q$-CP~\cite{Riccio2018} to use deep learning and improve both the
focused exploration and the generalization of the algorithm. Thanks to
the better generalization capabilities, \dop{} can be used in domains
with high-dimensional states, such as multi-agent scenarios. For this
reason, we evaluated both the representation and the algorithm on
three different tasks that involve multiple or hyper-redundant
robots. To better analyze the quality of the adopted representation,
we also transferred a learned policy from simulation to real-world.

Unfortunately, \dop{} still needs a predefined simulation
environment. This is not always available, and does not properly
capture the dynamics of the world, making our algorithm less appealing
in highly interactive scenarios. To address this issue, we aim at
learning the dynamics of the world at robot operation time, and
simultaneously improving its policy based on the learned
model~\cite{Weber2017}.



\balance
\bibliographystyle{ACM-Reference-Format}  
\bibliography{refs}  

\end{document}